\documentclass[letterpaper,10pt,conference]{ieeeconf}
\ifx\anonymous\undefined
\IEEEoverridecommandlockouts  
\fi
\usepackage{multicol}
\usepackage[bookmarks=true]{hyperref}
\usepackage{lineno}

\usepackage{booktabs}
\usepackage{color}
\usepackage{multirow}
\usepackage{xcolor}
\usepackage{graphicx}
\usepackage{colortbl}
\usepackage{amsmath}
\usepackage{amssymb}
\usepackage{siunitx}
\newcommand{\sys}{MVLAD-AD}
\title{\LARGE \bf Efficient and Explainable End-to-End Autonomous Driving via Masked Vision-Language-Action Diffusion}
\author{}
\ifx\anonymous\undefined 
\author{Jiaru Zhang, Manav Gagvani, Can Cui, Juntong Peng, Ruqi Zhang, and Ziran Wang
\thanks{J. Zhang, M. Gagvani and C. Cui contributed equally to this work.
J. Zhang is with the Institute for Physical Artificial Intelligence (IPAI), Purdue University, West Lafayette, IN 47907, USA.
M. Gagvani, C. Cui, J. Peng, and Z. Wang are with the College of Engineering, Purdue University, West Lafayette, IN 47907, USA.
R. Zhang is with the Department of Computer Science, Purdue University, West Lafayette, IN 47907, USA.
Corresponding author: Jiaru Zhang, e-mail: jiaru@purdue.edu.}
}
\fi
\begin{document}

\maketitle
\begin{abstract}
Large Language Models (LLMs) and Vision-Language Models (VLMs) have emerged as promising candidates for end-to-end autonomous driving. However, these models typically face challenges in inference latency, action precision, and explainability. 
Existing autoregressive approaches struggle with slow token-by-token generation, while prior diffusion-based planners often rely on verbose, general-purpose language tokens that lack explicit geometric structure.
In this work, we propose Masked Vision-Language-Action Diffusion for Autonomous Driving (\sys), a novel framework designed to bridge the gap between efficient planning and semantic explainability via a masked vision-language-action diffusion model. 
Unlike methods that force actions into the language space, we introduce a discrete action tokenization strategy that constructs a compact codebook of kinematically feasible waypoints from real-world driving distributions.
Moreover, we propose geometry-aware embedding learning to ensure that embeddings in the latent space approximate physical geometric metrics.
Finally, an action-priority decoding strategy is introduced to prioritize trajectory generation. 
Extensive experiments on nuScenes and derived benchmarks demonstrate that \sys~achieves superior efficiency and outperforms state-of-the-art autoregressive and diffusion baselines in planning precision, while providing high-fidelity and explainable reasoning.
\end{abstract}

\IEEEpeerreviewmaketitle

\section{Introduction}
The paradigm of autonomous driving is shifting from modular pipelines to end-to-end learning systems, where a unified model directly maps raw sensor inputs to driving decisions. 
However, traditional end-to-end models often function as black boxes, suffering from limited explainability and poor generalization in complex scenarios.
Recently, Large Language Models (LLMs) and Vision-Language Models (VLMs) have emerged as promising candidates for autonomous driving, enabling reasoning over complex traffic scenarios and human interaction.
By formulating driving as a language modeling problem, these models leverage rich world knowledge from pretraining and improve autonomous driving performance~\cite{mao2023gpt,wendilu,xu2024drivegpt4}.

\begin{figure}
    \centering
    \includegraphics[width=\linewidth]{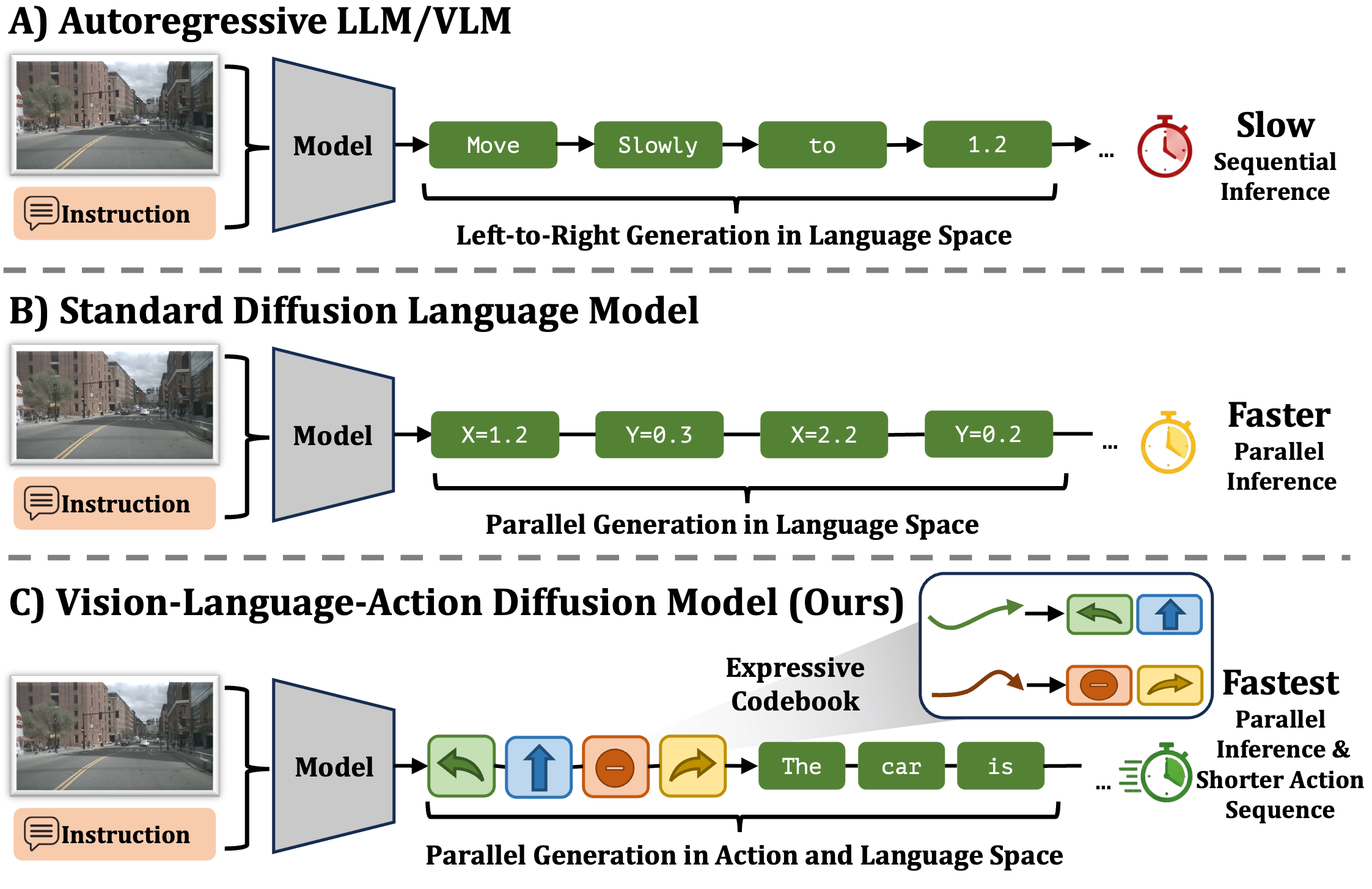}
    \caption{Motivation. Comparison of paradigms for LLM/VLM-based autonomous driving. (A) Autoregressive VLMs suffer from high latency due to token-by-token sequential generation. (B) Standard diffusion language models enable parallel generation but operate in a verbose language space. (C) Our proposed Vision-Language-Action Diffusion Model incorporates an expressive codebook to map continuous actions into compact discrete tokens. This design enables simultaneous parallel generation in both action and language spaces, significantly reducing sequence length and achieving the fastest inference speed.}
    \label{fig:motivation}
    \vspace{-0.1in}
\end{figure}

Despite their promise, current LLM/VLM models for autonomous driving face three main challenges: inference latency, action precision, and explainability.
Most existing approaches rely on autoregressive generation. 
While powerful, as illustrated in Figure~\ref{fig:motivation}(A), token-by-token inference is prohibitively slow for the latency-critical nature of autonomous driving.
Moreover, processing continuous actions in the language space results in verbose token representations, where describing precise trajectories requires excessive sequence lengths, limiting the inference efficiency of the planning framework.
Finally, existing models often struggle to produce plans together with coherent reasoning. Reliance on separate, post-hoc explanation modules often fails to align semantic reasoning with driving actions.

Diffusion language models have emerged as a powerful non-autoregressive alternative, enabling parallel decoding to address the inference latency bottleneck~\cite{nie2025large}.
In autonomous driving, ViLaD~\cite{cui2025vilad} pioneered this paradigm for efficient planning and achieved substantial improvements in both planning accuracy and inference speed. However, as shown in Figure~\ref{fig:motivation}(B), ViLaD relies on verbose, general-purpose language tokens to represent continuous trajectories, introducing representation redundancy.
Moreover, it only predicts the driving decisions without any semantic reasoning explanations.
These limit both the planning performance and the transparency of current diffusion language model-based planners.

Concurrently, Vision-Language-Action (VLA) models~\cite{kimopenvla,zitkovich2023rt} have demonstrated remarkable success by unifying perception, reasoning, and control within a single backbone for embodied AI.
By integrating control signals as specialized tokens within the language vocabulary, VLAs facilitate deep semantic grounding, enabling the alignment of physical maneuvers with high-level linguistic reasoning.
However, standard VLA architectures are predominantly tailored for robotic manipulation tasks, and the best way to leverage VLA modeling for end-to-end autonomous driving remains an open question.
Naively discretizing the driving action space results in an explosion of action tokens, rendering the search space intractable and hindering efficient learning.

In this work, we propose Masked Vision-Language-Action Diffusion for Autonomous Driving (\sys), a novel framework designed to achieve both efficient planning and semantic explainability.
Compared with approaches that force continuous actions into verbose language spaces, we introduce a discrete action tokenization strategy that constructs a compact codebook of kinematically feasible waypoints derived from real-world driving distributions, effectively compressing the action search space. 
We further propose geometry-aware embedding learning, which aligns the latent embedding space with geometric metrics. 
These components are integrated into a masked VLA diffusion transformer to model the joint probability of driving actions and linguistic explanations. 
To resolve the conflict between latency and explainability, we utilize an action-priority decoding strategy that prioritizes trajectory generation during inference. 
Extensive experiments on the nuScenes and derived reasoning datasets, including Nu-X and nuScenes-QA, demonstrate that \sys~achieves strong planning performance among VLM-based planners while providing high-fidelity, explainable reasoning.

In summary, our contributions are as follows: 
\begin{itemize}
\item We propose \sys, a novel end-to-end masked VLA diffusion framework that achieves highly efficient end-to-end autonomous driving while retaining semantic reasoning ability.
\item To bridge the modality gap, we propose discrete action tokenization to map trajectories to compact action tokens, and geometry-aware embedding learning to enforce metric consistency in the embedding space. We further introduce an action-priority decoding strategy to enable low-latency planning.
\item Extensive experiments demonstrate that \sys~outperforms baseline methods on the nuScenes planning benchmark while maintaining superior inference speed and generating coherent, physically aligned linguistic explanations. 
\ifx\anonymous\undefined
\else
The code and checkpoints will be released upon acceptance.
\fi
\end{itemize}

\section{Related Work}
\subsection{Large Language Model-Based End-to-End Autonomous Driving}
End-to-end systems are a newly emerging paradigm in autonomous driving that directly maps sensory inputs and ego-vehicle states to planning trajectories and/or low-level control actions~\cite{chen2024end}. Motivated by the remarkable progress of LLMs, recent studies have explored generative and autoregressive formulations for end-to-end autonomous driving. Chen et al.~\cite{chen2024drivinggpt} adopted an autoregressive generation strategy to sequentially predict future scene representations along with corresponding control actions. Following this paradigm, Huang et al.~\cite{huang2024drivegpt} further extended autoregressive modeling to a GPT-style architecture, where future driving scenes are represented and predicted as token sequences.

With the rapid advancement of multimodal foundation models (typically VLMs), their adoption in end-to-end autonomous driving accelerated significantly. Early efforts demonstrated the feasibility of directly incorporating language-centric models into closed-loop driving systems. Shao et al.~\cite{shao_lmdrive_2023} presented one of the first multimodal LLM-driven end-to-end driving frameworks operating in closed-loop settings
. Their system takes navigation goals, multimodal sensory observations, and auxiliary textual instructions as input and directly outputs low-level control commands. Similarly, Wang et al.~\cite{wang_drivemlm_2023} explored the integration of LLMs into autonomous driving pipelines by aligning language-based decision-making with high-level vehicle control, enabling multimodal inputs and providing interpretable decision rationales.

Beyond direct control generation, a parallel research direction focuses on exploiting the reasoning capabilities of foundation models for decision support and scene understanding. Xu et al.~\cite{xu2024drivegpt4} introduced a question–answering-based driving framework that formulates driving decisions as structured queries. Extending this, Sima et al.~\cite{sima_drivelm_2024} proposed a graph-based visual question answering (VQA) paradigm, where logically dependent QA pairs capture complex relational reasoning in traffic scenarios. To further enhance explicit reasoning supervision, Wang et al.~\cite{wang_drivecot_2024} constructed a chain-of-thought (CoT) driving dataset, annotating both intermediate reasoning steps and final decisions, and proposed a baseline that injects multiple task-specific learnable queries into the reasoning process. More comprehensive multimodal reasoning systems were later introduced by Hwang et al.~\cite{hwang2024emma} and Xing et al.~\cite{xing2025openemma}, which demonstrated strong generalization, robustness, and scalability across diverse 
driving environments. However, most of these methods are built upon the autoregressive generation paradigm, resulting in relatively slow inference due to sequential decoding and a left-to-right generation pattern, while primarily modeling short-term dynamics by predicting only the next-step actions or frames.

\subsection{Explainable Autonomous Driving}
Explainability has long been recognized as a critical challenge for learning-based autonomous driving systems. As deep neural networks are increasingly adopted for perception, decision-making, and control, their ``black box" nature makes it difficult to understand why specific driving actions are produced. This lack of transparency directly affects system verification, safety assurance, trust building, and human–machine collaboration, motivating extensive research on explainable artificial intelligence (XAI) methods for autonomous driving~\cite{arrieta2020explainable,omeiza2021explanations}.

A broad class of explainability approaches focuses on model-agnostic explanation techniques~\cite{lundberg2017unified,shrikumar2017learning,ribeiro2016should}. These methods aim to explain individual predictions by estimating feature relevance or attribution without requiring access to the internal structure of the model. In parallel, model-specific techniques have been proposed to analyze internal representations, including gradient-based attribution~\cite{selvaraju2017grad}, saliency visualization~\cite{simonyan2013deep}, and attention-based explanations~\cite{xu2015show}.

Within the autonomous driving domain, attention mechanisms have been widely adopted to visualize spatial regions in driving scenes that have a strong influence on control decisions~\cite{kim2017interpretable}. To further enhance explainability, attention-based driving models have been combined with natural language generation, enabling systems to output both control actions and corresponding textual explanations grounded in visual observations~\cite{kim2018textual}. This line of work was later extended by incorporating linguistic structure and decoding constraints, leading to more coherent and informative textual commentaries for driving behavior~\cite{kuhn2023textual}. However, attention-based explanations alone have been shown to be insufficient for faithfully reflecting model reasoning~\cite{jain2019attention}. This limitation motivated hybrid explainability strategies that jointly leverage attention weights, encoded features, gradients, and class activation signals to better approximate the underlying decision logic of transformer-based architectures~\cite{qiang2022attcat}.

Recently, LLMs and VLMs introduced a paradigm shift in explainability for autonomous driving, using language as an explicit medium for expressing perception, reasoning, and decision rationales. For example, language-driven driving systems such as DriveLM~\cite{sima_drivelm_2024} and LLM-based closed-loop driving frameworks~\cite{shao_lmdrive_2023, wang_drivemlm_2023, chen2024driving, 10919978, cui2025llm4adlargelanguagemodels} provide textual explanations that articulate what the vehicle observes and why specific actions are taken. Question--answering--based formulations further enable interactive explainability by allowing models to respond to structured queries about driving scenes and decisions~\cite{xu2024drivegpt4}. More comprehensive multimodal reasoning frameworks, including EMMA~\cite{hwang2024emma} and OpenEMMA~\cite{xing2025openemma}, demonstrate that language-centric explanations can generalize across diverse driving scenarios while improving robustness and transparency.

\section{Methodology}
\begin{figure*}
    \centering
    \includegraphics[width=\linewidth]{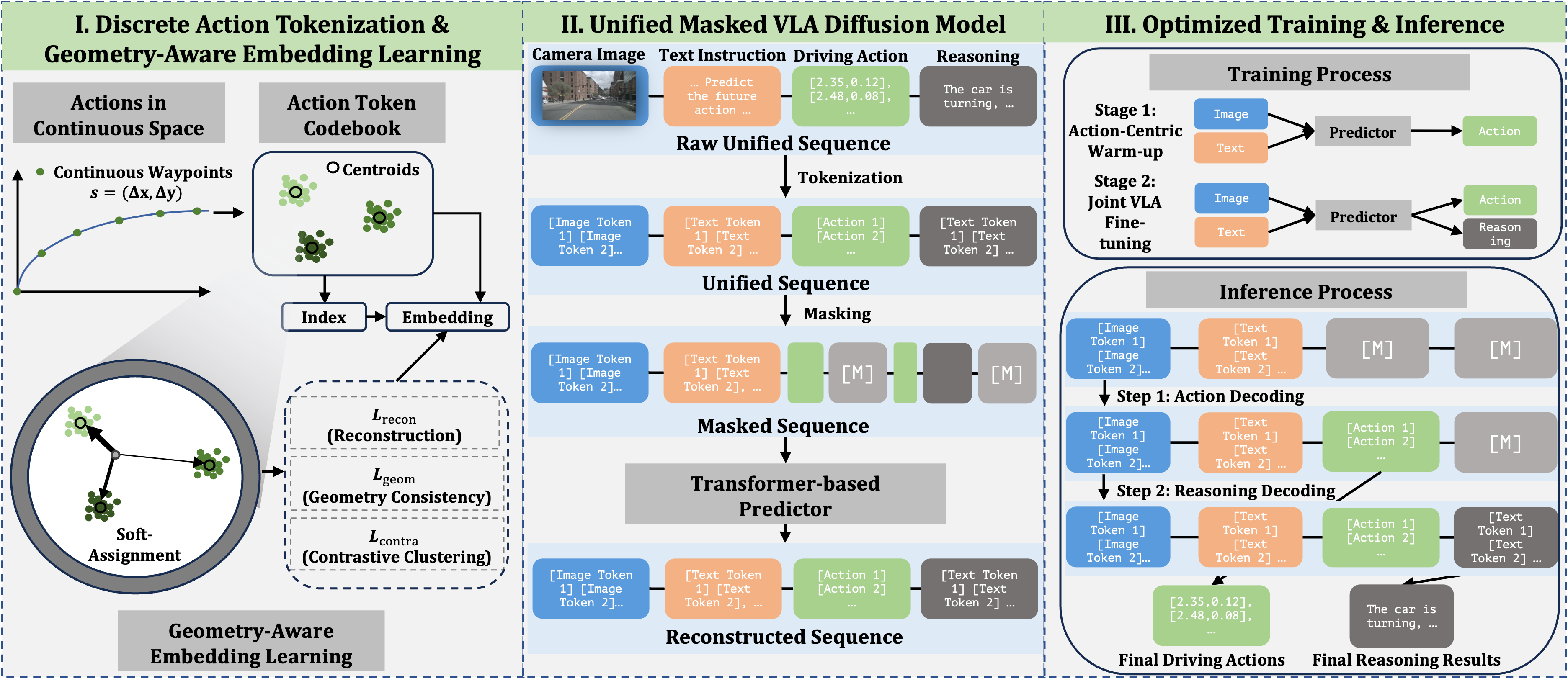}
    \caption{An overview of \sys. (A) Discrete Action Tokenization and Geometry-Aware Embedding Learning: We construct a compact codebook of driving actions from real-world data and learn a geometry-aware embedding space via soft-assignment and geometric consistency objectives. (B) Unified Masked VLA Diffusion: Visual, instruction, action, and reasoning tokens are unified into a single sequence for masked generative modeling. (C) Optimized Training \& Inference: During training, we employ a two-stage learning strategy to assist training. During inference, an action-priority decoding strategy is introduced to prioritize trajectory generation for low latency, ensuring the reasoning explanation is highly faithful to the driving actions.}
    \label{fig:overview}
        \vspace{-0.05in}
\end{figure*}

\subsection{Framework Overview}
\label{sec:framework}
As illustrated in Figure~\ref{fig:overview}, \sys~formulates end-to-end autonomous driving as a \textit{conditional masked generative modeling} problem.
We construct a unified sequence of discrete tokens, $\mathbf{x} = [\mathbf{x}^c; \mathbf{x}^g]$, from heterogeneous modalities.
The conditioning part, $\mathbf{x}^c$, aggregates multi-view visual tokens $\mathbf{x}^v$ and textual instruction tokens $\mathbf{x}^i$.
The target generation part, $\mathbf{x}^g$, concatenates discrete action tokens $\mathbf{x}^a$ representing the future trajectory and reasoning tokens $\mathbf{x}^r$ explaining the decision.
Our objective is to learn the conditional distribution $p_\theta(\mathbf{x}^g \mid \mathbf{x}^c)$ by reversing a parallel masking process~\cite{nie2025large}.

\noindent\textbf{Multimodal Input Encoding.}
To process the heterogeneous inputs, we employ specialized encoders to project all signals into a shared embedding space, ensuring semantic alignment across modalities:
\begin{itemize}
    \item \textbf{Vision:} Multi-view camera images are encoded by a pre-trained vision encoder and mapped to the transformer's dimension via a learnable MLP Projector.
    \item \textbf{Instruction:} The text instruction is tokenized using a standard text tokenizer and converted into an embedding to condition the model on a specific driving task.
    \item \textbf{Action:} We use our proposed Discrete Action Tokenizer detailed in Section~\ref{sec:action_token} to map continuous waypoints into discrete action tokens, and use our proposed Geometry-Aware Embedding Learning detailed in Section~\ref{sec:embeddings} to obtain their embeddings. 
    \item \textbf{Reasoning:} Similar to the instruction, the reasoning is tokenized and embedded as text.
\end{itemize}

\noindent\textbf{Unified Sequence Modeling.}
Unlike modular pipelines that isolate planning from explanation, \sys~concatenates these embeddings into a single sequence, $\mathbf{X} = [\mathbf{X}^v; \mathbf{X}^i; \mathbf{X}^a; \mathbf{X}^r]$, where $\mathbf{X}^v$, $\mathbf{X}^i$, $\mathbf{X}^a$, and $\mathbf{X}^r$ correspond to the embeddings of visual features, textual instructions, action tokens, and reasoning tokens, respectively.
This unified representation enables the model to leverage bidirectional attention mechanisms and capture the interdependencies among visual observations, text instructions, vehicle actions, and semantic reasoning in a global context.

\noindent\textbf{Masked Generative Transformer.} 
The core of \sys~is a Transformer-based predictor for masked diffusion generative modeling. 
As detailed in Section~\ref{sec:training}, with a training strategy that jointly learns action and reasoning, the transformer takes the unified sequence as input, with parts of $\mathbf{x}^a$ and $\mathbf{x}^r$ masked, and learns to reconstruct the masked tokens.
This architecture effectively models the joint distribution $p_\theta(\mathbf{x}^a, \mathbf{x}^r \mid \mathbf{x}^v, \mathbf{x}^i)$, providing the flexibility to support both efficient parallel planning and reasoning generation, as detailed in Section~\ref{sec:inference}.

\subsection{Driving Action Tokenization} \label{sec:action_token} 
To bridge the modality gap between continuous trajectory planning and discrete language generation, we propose a Discrete Action Tokenization strategy. 
In this strategy, the trajectories are modeled as sequences of discrete driving action tokens. 
Concretely, we represent the trajectory as a series of waypoints in the local ego-coordinate system. 
Let $\mathbf{w} = (x, y) \in \mathbb{R}^2$ denote a single waypoint representing the vehicle's position relative to the ego-vehicle at a specific future timestamp, and
let $\mathcal{D} = \{\mathbf{w}_1, \mathbf{w}_2, \dots, \mathbf{w}_M\}$ represent a set of all valid waypoints collected from the real-world driving dataset, where $M$ is the total number of waypoints in the training corpus across all time horizons. Our objective is to construct a compact codebook $\mathcal{C} = \{\mathbf{c}_1, \mathbf{c}_2, \dots, \mathbf{c}_N\}$ consisting of $N$ representative waypoints, such that any continuous waypoint $\mathbf{w} \in \mathcal{D}$ can be approximated by a representative centroid $\mathbf{c} \in \mathcal{C}$ with minimal error. This is formulated as an optimization problem to minimize the within-cluster sum of squares:
\begin{align} 
\mathcal{J} = \sum_{i=1}^{M} \min_{\mathbf{c} \in \mathcal{C}} \| \mathbf{w}_i - \mathbf{c} \|_2^2, \label{eq:kmeans}
\end{align}
where $\| \cdot \|_2$ denotes the Euclidean distance. Utilizing a standard K-Means solver, we obtain the optimal set of spatial centroids $\mathcal{C}$, which serve as our discrete action tokens.

During training and inference, a quantizer $Q(\cdot)$ is employed to map any continuous predicted waypoint $\hat{\mathbf{w}}$ to the index of its nearest centroid in the codebook:
\begin{align}
k = \arg\min_{j \in \{1, \dots, N\}} \| \hat{\mathbf{w}} - \mathbf{c}_j \|_2.
\end{align}
This discretization transforms the trajectory generation problem into a sequence of $N$-way classification problems, constraining the output space to physically feasible spatial primitives.

\subsection{Geometry-Aware Embedding Learning}
\label{sec:embeddings} 
While the action tokenization provides a compact codebook of tokens, treating these tokens as independent categorical indices and using randomly initialized embeddings discards the rich metric information inherent in the trajectory space.
To enable the diffusion language model to reason about the physical properties of actions, we propose a pre-training stage to learn a geometry-aware embedding space.
Let $\mathbf{E} \in \mathbb{R}^{N \times D}$ denote the learnable embedding matrix for our $N$ action tokens. We aim to optimize $\mathbf{E}$ such that the Euclidean distance in the latent space approximates the geometric distance in the physical space.

\noindent\textbf{Soft-Assignment and Reconstruction.} 
To stabilize the optimization and bridge the gap between continuous inputs and discrete tokens, we employ a temperature-scaled soft-assignment mechanism during training. Given a ground-truth waypoint $\mathbf{w}$, instead of performing a hard lookup, we compute a weighted embedding $\mathbf{z}$ based on the distances to the top-$K$ nearest centroids in the codebook $\mathcal{C}$:
\begin{equation}
\mathbf{z} = \sum_{j \in \mathcal{N}_K(\mathbf{w})} \frac{\exp(-\|\mathbf{w} - \mathbf{c}_j\|_2 / \tau)}{\sum_{l \in \mathcal{N}_K(\mathbf{w})} \exp(-\|\mathbf{w} - \mathbf{c}_l\|_2 / \tau)} \mathbf{E}_j,
\end{equation}
where $\mathcal{N}_K(\mathbf{w})$ denotes the set of indices of the $K$ nearest centroids, and $\tau$ is a temperature parameter.
We then use a lightweight MLP decoder $D_\phi: \mathbb{R}^D \to \mathbb{R}^2$ to reconstruct the original coordinates, minimizing the reconstruction loss $\mathcal{L}_{\text{recon}} = \| D_\phi(\mathbf{z}) - \mathbf{w} \|_2^2$.

\noindent\textbf{Metric Alignment Objectives.} 
To explicitly enforce the geometric structure, we introduce two auxiliary losses: 
\begin{enumerate}
    \item \textbf{Geometry Consistency Loss.} We enforce that pairwise distances in the embedding space correlate with physical distances. For a batch of pairs $(\mathbf{w}_i, \mathbf{w}_j)$, we minimize the discrepancy between their normalized distances:
    \begin{equation}
    \mathcal{L}_{\text{geom}} = \mathbb{E}_{i,j} \left[ \left( \frac{\| \mathbf{z}_i - \mathbf{z}_j \|_2}{\bar{d}_z} - \frac{\| \mathbf{w}_i - \mathbf{w}_j \|_2}{\bar{d}_w} \right)^2 \right],
    \end{equation}
    where $\bar{d}_z$ and $\bar{d}_w$ are the median pairwise distances in the batch, serving as robust scaling factors.
    
    \item \textbf{Contrastive Clustering Loss.} We apply a supervised contrastive loss to structure the latent space. For an anchor point $i$ with normalized embedding $\tilde{\mathbf{z}}_i$, let $P(i)$ be the set of other points in the batch assigned to the same centroid index. The loss pulls positive pairs together while pushing apart negatives:
    \begin{align}
        \begin{aligned}
                \mathcal{L}_{\text{contra}} = \sum_{i \in \mathcal{B}} \frac{-1}{|P(i)|} \sum_{p \in P(i)} \log \frac{\exp(\tilde{\mathbf{z}}_i \cdot \tilde{\mathbf{z}}_p / \tau_{\text{con}})}{\sum\limits_{a \in A(i)} \exp(\tilde{\mathbf{z}}_i \cdot \tilde{\mathbf{z}}_a / \tau_{\text{con}})},
        \end{aligned}
    \end{align}
    where $A(i)$ is the set of all other indices in the batch, and $\tau_{\text{con}}$ is a contrastive temperature.
\end{enumerate}

\noindent\textbf{Optimization and Curriculum.}
The final objective is a weighted sum of the three losses.
We employ a curriculum learning strategy to ease the transition from continuous to discrete representation. The parameter $K$ decays from 16 to 1 over the course of training. This allows the model to capture the local manifold structure via soft interpolation in the early stages, eventually converging to a hard discrete mapping that yields the embedding matrix $\mathbf{E}$.

\subsection{Training Process}\label{sec:training} 
Training a unified VLA model involves learning two distinct capabilities of precise planning and high-level semantic reasoning. 
Following the standard masked diffusion formulation~\cite{nie2025large}, the base training objective is the negative log-likelihood computed on the masked tokens:
\begin{align}
    \mathcal{L}_{\text{diff}} = \mathbb{E}_{t, \mathbf{x}_0} \left[ - \sum_{i \in \mathcal{M}_t} \log p_\theta(x_{0,i} | \mathbf{x}_t) \right], \label{equ:diff_loss}
\end{align}
where $\mathcal{M}_t$ denotes the set of masked indices at step $t$.
Since the discrete action tokens introduced in Section~\ref{sec:action_token} represent a new modality with no pre-trained knowledge base, learning their distributions is more challenging than standard language modeling. To address this, we use a two-stage learning strategy.

\noindent\textbf{Stage 1: Action-Centric Warm-up.}
In the first stage, we isolate the trajectory generation task to initialize the learning of the novel action modality.
Concretely, we exclude the reasoning token subsequence $\mathbf{X}^r$ from the input sequence entirely, constructing a shortened sequence $\mathbf{X}_{\text{stage1}} = [\mathbf{X}^v; \mathbf{X}^i; \mathbf{X}^a]$.
We mask a portion of the action-token indices and train the model to reconstruct them conditioned solely on visual and instructional contexts.
This focused objective forces the model to learn about the physical dynamics and the geometric structure of the action codebook, establishing a robust motion prior without the distraction of the language generation task.

\noindent\textbf{Stage 2: Joint VLA Fine-tuning.} In the second stage, we introduce the reasoning task where the unified sequence contains both the action-token subsequence $\mathbf{X}^a$ and the reasoning-token subsequence $\mathbf{X}^r$. We mask both action-token indices and reasoning-token indices, training the model to generate the full output sequence. This stage aligns the physical actions with semantic explanations, leveraging the motion priors established in Stage 1.

\subsection{Inference Process}
\label{sec:inference}

To balance the conflicting requirements of low-latency planning and semantic explainability, we utilize an action-priority decoding strategy. 
Standard masked diffusion methods typically employ a global confidence-based sampling schedule, where the tokens with the highest confidence scores across the \textit{entire} sequence are unmasked first. However, this unconstrained approach might generate text tokens and action tokens in a mixed order, delaying the availability of actions.

Instead, we enforce a modality-constrained unmasking policy that prioritizes the resolution of the trajectory.
Let $\mathbf{x}_t$ denote the sequence at diffusion step $t$, containing both action-token positions and reasoning-token positions. In each iteration, the transformer predictor outputs the probability distribution  $p_\theta(\mathbf{x}_0 \mid \mathbf{x}_t)$ for \textit{all} masked tokens.
Although the model predicts the entire sequence simultaneously, we restrict the unmasking candidate set exclusively to the action indices until planning is fully determined.
Specifically, we compute the confidence score $u_j = \max_{k} p_\theta(x_j = k \mid \mathbf{x}_t)$ for all masked action-token positions $j$. We then select the subset of action tokens with the highest confidence scores to demask, i.e., replace $[\texttt{M}]$ with the predicted token ID, while keeping all reasoning tokens in the masked state.

This strategy yields two critical benefits. By focusing the decoding budget solely on the $N_a$ action tokens (where $N_a$ is much smaller than the text length) in the initial steps, the trajectory is finalized and ready for execution significantly faster than waiting for the full sequence to converge. On the other hand, once the action tokens are fully unmasked, they serve as fixed, fully-observed conditions for the subsequent generation of reasoning tokens. This implicitly enforces semantic consistency, as the generated explanation is conditioned on a deterministic future plan.

\section{Experiments}
We evaluate the proposed \sys~framework across both efficient planning and linguistic reasoning. To ensure a comprehensive comparison, we compare against a wide range of baselines, including both open-source state-of-the-art models and representative commercial large language models.

\noindent\textbf{Datasets and Metrics.}
We primarily utilize the nuScenes dataset \cite{caesar2020nuscenes} for planning evaluation. 
Specifically, we conduct all experiments and baseline comparisons under a single-frame setting for efficiency.
Following standard protocols, we report the L2 displacement error at \SI{1}{\second}, \SI{2}{\second}, and \SI{3}{\second} horizons, along with the average L2 error. For the reasoning tasks, we leverage two specialized datasets: Nu-X \cite{dinghint}, a dataset focused on explaining driving decisions, and nuScenes-QA \cite{qian2024nuscenes}, a large-scale visual question-answering benchmark for autonomous driving scenarios.
For Nu-X, we employ standard natural language generation metrics, including BLEU-4, METEOR, ROUGE-L, and CIDEr, to measure the semantic alignment between generated explanations and ground truths. 
For nuScenes-QA, following the literature \cite{dinghint,ma2025aln}, we report accuracy across zero-hop (H0), one-hop (H1), and overall questions (All).

\noindent\textbf{Baselines.}
We categorize our baselines based on the task and model architecture:

\textit{1) Planning Baselines:} We compare \sys~with a series of representative VLM-based end-to-end driving methods. For autoregressive VLMs, we include models fine-tuned on different backbones: LLaVA-1.6-Mistral-7B, Llama-3.2-11B-Vision-Instruct, and Qwen2-VL-7B-Instruct. We also compare against the best models from the DriveLM \cite{sima_drivelm_2024}, OpenEMMA \cite{xing2025openemma}, and LightEMMA \cite{qiao2025lightemma} frameworks. For diffusion VLMs, we include ViLaD \cite{cui2025vilad}, which serves as the direct baseline for demonstrating the advantages of our joint VLA modeling and discrete action tokenization. Additionally, we evaluate against UniAD \cite{hu2023planning}, a representative end-to-end autonomous driving model. For a fair comparison, the reported UniAD results are derived from its single-frame version, as evaluated in \cite{sima_drivelm_2024}.

\textit{2) Explanation Baselines:} For reasoning tasks, we compare with a series of specialized models, including Hint-AD \cite{dinghint}, ALN-P3 \cite{ma2025aln}, and TOD3Cap \cite{jin2024tod3cap}, which are specifically designed for driving reasoning. We also benchmark against some representative closed-source foundation models, GPT-4o and Gemini-1.5, to estimate how our specialized 7B model compares against general models in industry.

\noindent\textbf{Implementation Details.}
\sys~is implemented using the PyTorch framework.
We initialize our model weights from the pre-trained LLaDA checkpoint \cite{nie2025large}.
To improve parameter efficiency, we employ Low-Rank Adaptation (LoRA) with a rank of $r=256$.
Training is distributed across 4 NVIDIA H100 GPUs using \texttt{bfloat16} precision, with a global batch size of 32.
Each training stage consists of 8 epochs, requiring approximately \SI{9}{\hour} to complete.
For evaluation, all inference experiments are conducted on a single NVIDIA A100 GPU.

\begin{table}[t]
    \centering
    \caption{Planning Errors (\si{\meter}) at Different Horizons and Planning Failure Rate (FR). Dashes indicate that the method does not report the corresponding results.}
        \label{tab:fdes}
    \begin{tabular}{l|cccc|c}
        \toprule
        Method & \SI{1}{\second} ($\downarrow$) & \SI{2}{\second} ($\downarrow$) & \SI{3}{\second} ($\downarrow$) & Avg ($\downarrow$) & FR ($\downarrow$)\\
        \midrule
                LLaVA-1.6 
        & 0.91 & 2.50 & 3.44 & 2.28 & 55.25\% \\
        Llama-3.2
        & 0.80 & 2.31 & 3.10 & 2.07 &0.06\% \\
        Qwen2-VL  
        & 1.32 & 2.94 & 3.98 & 2.74 & 0.03\% \\
        OpenEMMA
        & 1.45 &3.21 &3.76 &2.81 & 16.11\% \\
        LightEMMA & - & - & 2.90 & 1.45 & \textbf{0.00\%} \\
        UniAD-Single & - & - & - & 1.80 & - \\
        DriveLM & - & - & - & 1.39 & - \\
    
         ViLaD
        & 0.81 & 1.93 & 2.69 & 1.81  & \textbf{0.00\%} \\ \midrule
        \sys & \textbf{0.70} & \textbf{1.31} & \textbf{2.34} &   \textbf{1.28} &  \textbf{0.00\%} \\ 
        \bottomrule
    \end{tabular}
    \vspace{-0.1in}
\end{table}
\subsection{Planning Comparison}
\noindent\textbf{Superiority of Diffusion Language Modeling.} First, we compare diffusion-based architectures, \sys~and ViLaD, against standard autoregressive VLM baselines. As shown in Table~\ref{tab:fdes}, diffusion-based methods consistently outperform other baselines across all time horizons. Specifically, \sys~achieves an average L2 error of \SI{1.28}{\meter}, significantly reducing the error compared to the baselines. We attribute this gap to the inherent advantage of diffusion modeling for the planning task, where diffusion-based approaches (\sys~and ViLaD) allow the model to better capture the nature of driving behaviors without the limitations of sequential prediction.

\noindent\textbf{Benefits of VLA Modeling.} Within the diffusion framework, \sys~further surpasses the previous state-of-the-art method ViLaD, reducing the average L2 error from \SI{1.81}{\meter} to \SI{1.28}{\meter}. This performance gain verifies the effectiveness of the discrete action tokenization. Unlike ViLaD, which operates only in language space, our framework projects the action space into a compact codebook of $N=256$ representative action tokens. Therefore, \sys~performs a classification task over a set of driving actions learned from real-world data. This reduces the complexity of the prediction space, allowing the model to lock onto the optimal trajectory with higher precision and stability.

\noindent\textbf{Systematic Robustness.} Reliability is a prerequisite for robotic deployment. We observe that general-purpose VLMs like LLaVA-1.6 suffer from a high failure rate (55.25\%), largely due to format hallucinations where the model fails to adhere to the rigid output syntax required for planning. In contrast, \sys~demonstrates a 0.00\% failure rate. By mapping the action space to a fixed codebook of valid trajectory centroids and employing a structured decoding process, our method structurally guarantees that every generated output corresponds to a valid trajectory. This structural constraint eliminates the risk of format errors inherent in text-based planners, ensuring the system's robustness for downstream execution.

\begin{figure}
    \centering
    \includegraphics[width=\linewidth]{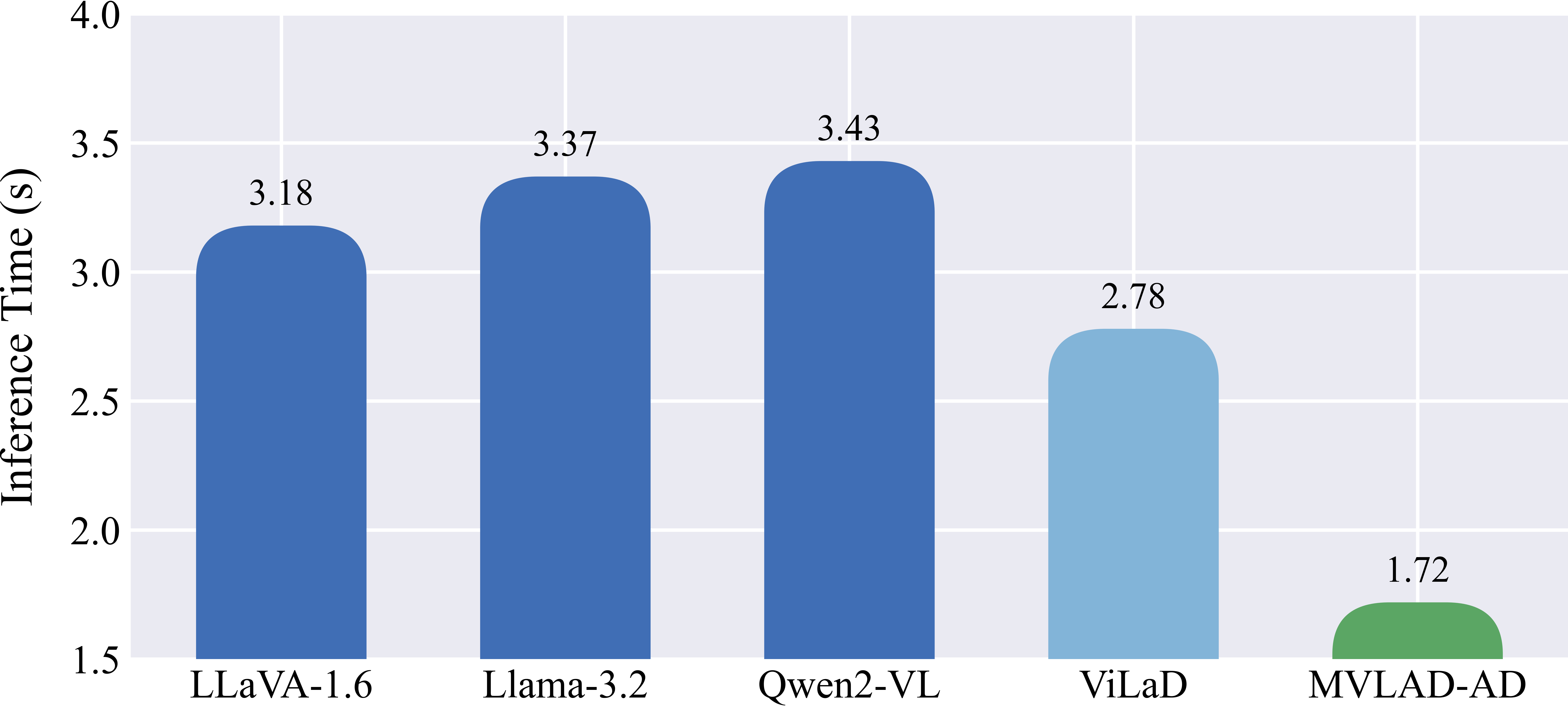}
    \caption{Planning inference time comparison among autoregressive and diffusion-based methods on a single NVIDIA A100 GPU.}
    \label{fig:time_comparison}
\end{figure}

\noindent\textbf{Inference Efficiency.} Figure~\ref{fig:time_comparison} presents the inference latency comparison. Diffusion-based architectures inherently outperform autoregressive baselines by leveraging parallel decoding. \sys~achieves a further acceleration by introducing a compact VLA modeling scheme.
Instead of predicting trajectories via redundant text tokens, our discrete action tokenization compresses complex motion into a minimal sequence of motion primitives. This effectively shortens the sequence length required for planning, significantly reducing the computational workload for the diffusion denoiser.
Consequently, \sys~achieves an inference time of \SI{1.72}{\second}. This represents a $1.6\times$ speedup over the diffusion baseline ViLaD, and a $1.84\times$ speedup over LLaVA-1.6, establishing \sys~as an efficient solution for end-to-end autonomous driving.


\subsection{Reasoning Comparison}
\begin{table}[t]
\centering
\caption{Performance Comparison on Nu-X}
\label{tab:nux_results}
\resizebox{\linewidth}{!}{
\begin{tabular}{c|cccc}
\toprule
\multirow{2}{*}{Method} & \multicolumn{4}{c}{Nu-X} \\
 & CIDEr ($\uparrow$) & BLEU-4 ($\uparrow$) & METEOR ($\uparrow$)& ROUGE-L ($\uparrow$)\\ \midrule
TOD3Cap     & 14.5 & 2.45 & 10.5 & 23.0 \\
GPT-4o      & 19.0 & 3.95 & 10.3 & 24.9 \\
Gemini-1.5  & 17.6 & 3.43 & 9.3  & 23.4 \\
Hint-AD     & 22.4 & 4.18 & 13.2 & 27.6 \\
ALN-P3      & \textbf{28.6} & 5.59 & 14.7 & 35.2 \\ \midrule
\sys        & 19.5 & \textbf{13.0} & \textbf{36.8} & \textbf{37.3} \\ \bottomrule
\end{tabular}
}
\end{table}

\begin{table}[t]

\centering

\caption{Performance Comparison on nuScenes-QA}

\label{tab:nuscenes_qa_results}


\begin{tabular}{c|ccc}

\toprule

\multirow{2}{*}{Method} & \multicolumn{3}{c}{nuScenes-QA} \\

 & H0 ($\uparrow$)& H1 ($\uparrow$)& All ($\uparrow$)\\ \midrule

TOD3Cap     & 53.0 & 45.1 & 49.0 \\

GPT-4o      & 42.0 & 34.7 & 37.1 \\

Gemini-1.5  & 40.5 & 32.9 & 35.4 \\

Hint-AD     & 55.4 & 48.0 & 50.5 \\

ALN-P3      & 57.1 & 50.9 & 52.9 \\ \midrule

\sys        & \textbf{58.5}    & \textbf{54.3}    & \textbf{55.7}    \\ \bottomrule

\end{tabular}

\end{table}

\noindent\textbf{Driving Explanation on Nu-X.} Table~\ref{tab:nux_results} presents the quantitative results on the Nu-X dataset. \sys~demonstrates superior performance across most metrics compared to both general-purpose LVLMs, e.g., GPT-4o and Gemini-1.5, and specialized autonomous driving models, including Hint-AD and ALN-P3. Notably, our method achieves a BLEU-4 score of 13.0 and a METEOR score of 36.8, surpassing the previous state-of-the-art method ALN-P3 by a significant margin. Although ALN-P3 yields a higher CIDEr score, our substantial lead in BLEU-4 and ROUGE-L indicates that \sys~generates descriptions that are not only semantically richer but also more precisely aligned with reference captions in terms of n-gram overlap. This confirms that our model is able to effectively generate high-quality, coherent reasoning texts for complex driving scenarios.

\noindent\textbf{Visual Question Answering on nuScenes-QA.} Table~\ref{tab:nuscenes_qa_results} details the evaluation results on the nuScenes-QA benchmark. \sys~achieves an accuracy of 55.7\% on the overall split, consistently outperforming both large-scale commercial models and specialized driving agents. 
The performance leap indicates that \sys~captures complex dependencies in driving scenes, enabling the model to answer intricate questions about traffic dynamics with higher precision.

\subsection{Ablation Study}

\noindent\textbf{Impact of Action Vocabulary Size $N$.} 
The size of the discrete codebook $N$ introduces a fundamental trade-off between quantization precision and model learnability. 
Theoretically, a larger $N$ reduces the quantization error (the distance between a continuous waypoint and its nearest centroid), thereby raising the upper bound of reconstruction fidelity. 
However, an excessive number of tokens increases the complexity of the classification task.

We investigate this trade-off by training \sys~with vocabulary sizes $N \in \{128, 256, 384\}$, as reported in Table~\ref{tab:ablation-token-num}. We observe that $N=256$ strikes the optimal balance, achieving the lowest planning error of \SI{1.28}{\meter}. Crucially, increasing $N$ to 384 leads to a performance degradation to \SI{2.76}{\meter} with an obvious increase in final training loss from 0.36 to 0.53. This indicates that despite higher theoretical precision, the model struggles to converge due to the optimization difficulty in distinguishing between dense action tokens. Conversely, reducing $N$ to 128 yields the lowest training loss of 0.32, suggesting an easier classification task, but the planning error rises to \SI{1.73}{\meter}. This confirms that further reducing the codebook size creates a quantization bottleneck that limits physical precision, regardless of stable training convergence.

\begin{table}[ht]
\centering

\caption{Ablation on Action Vocabulary Size $N$}
\label{tab:ablation-token-num}
\resizebox{\linewidth}{!}{
\begin{tabular}{l|cc}
\toprule
Number of Action Tokens $N$ & Final Training Loss ($\downarrow$) & L2 (\si{\meter}) Avg. ($\downarrow$) \\ \midrule
128                  &  0.32     & 1.73            \\
256 (Default)         &   0.36  & \textbf{1.28}        \\
384       &     0.53  & 2.76 \\ \bottomrule
\end{tabular}
}   
\end{table}

\noindent\textbf{Effectiveness of Geometry-Aware Embedding Learning.}
We investigate the impact of geometry-aware embedding learning on the planning metrics of the model.
As a comparison, we remove this module and train the model with random embeddings. This causes a substantial performance degradation, increasing the average L2 error from \SI{1.28}{\meter} to \SI{2.39}{\meter}.
This comparison confirms that treating action tokens as independent categorical indices discards useful metric information, making it difficult for the model to learn effective planning.
By enforcing geometric consistency, our method ensures that the distance between tokens in the latent space correlates with their physical displacement, enabling the model to generate accurate trajectories.

\noindent\textbf{Action Representation: Waypoints vs. Displacements.} 
We analyze the impact of modeling trajectories as sequences of absolute waypoints versus relative displacements, which is another modeling paradigm adopted in the literature \cite{tang2026hermes}. 
As shown in Table~\ref{tab:ablation-modeling}, while the choice of representation has a marginal impact on pure planning performance, with the average L2 error degrading slightly from \SI{1.28}{\meter} to \SI{1.30}{\meter}, it causes a catastrophic collapse in reasoning capabilities.
Specifically, the displacement prediction model fails to generate coherent explanations, with the CIDEr score dropping to 0.08 and BLEU-4 decreasing by over 50\% compared to the waypoint model.
This confirms the rationale for applying waypoint prediction in our framework, as the parallel generation nature of diffusion struggles to implicitly integrate relative steps without explicit spatial anchors.

\begin{table}[ht]
\centering
\caption{Ablation on Action Representation. C: CIDEr, B: BLEU-4, M: METEOR, R: ROUGE-L.}
\label{tab:ablation-modeling}
\resizebox{\linewidth}{!}{
\begin{tabular}{l|c|cccc}
\toprule
\multirow{2}{*}{Modeling} & Planning & \multicolumn{4}{c}{Nu-X} \\
 & L2 (\si{\meter}) Avg. ($\downarrow$) & C ($\uparrow$)& B ($\uparrow$)& M ($\uparrow$)& R ($\uparrow$)\\ \midrule
Displacement & 1.30 & 0.08 & 5.66 & 23.1 & 26.4 \\
Waypoint & \textbf{1.28} & \textbf{19.5} & \textbf{13.0} & \textbf{36.8} & \textbf{37.3} \\ \bottomrule
\end{tabular}
}
\end{table}

\section{Conclusion}
\label{sec:conclusion}

In this work, we introduced \sys, a unified framework for end-to-end autonomous driving that balances low-latency, high-precision planning with semantic explainability. 
Instead of regular language space representations, our discrete action tokenization strategy successfully transforms continuous trajectory planning into a robust classification task over kinematically feasible motion primitives. 
Moreover, our geometry-aware embedding bridges the gap between semantic reasoning and physical dynamics, enabling the system to generate high-fidelity trajectories with physically grounded explanations. 
Our action-priority decoding strategy further reduces the planning latency in inference.
Extensive evaluations demonstrate that \sys~establishes strong performance on nuScenes planning and related language benchmarks, significantly outperforming autoregressive baselines in both planning precision and inference speed, while also providing high-quality explainable reasoning.

\ifx\anonymous\undefined
\section*{Acknowledgments}
This work used Anvil \cite{song2022anvil} at Purdue University through allocation CIS251316 from the Advanced Cyberinfrastructure Coordination Ecosystem: Services \& Support (ACCESS) program \cite{boerner2023access}, which is supported by National Science Foundation grants \#2138259, \#2138286, \#2138307, \#2137603, and \#2138296.
\fi
\bibliographystyle{IEEEtran}
\bibliography{main.bib}

@article{mao2023gpt,
  title={{GPT-Driver: Learning to Drive with GPT}},
  author={Mao, Jiageng and Qian, Yuxi and Ye, Junjie and Zhao, Hang and Wang, Yue},
  journal={arXiv preprint arXiv:2310.01415},
  year={2023}
}

@inproceedings{wendilu,
  title={{DiLu: A Knowledge-Driven Approach to Autonomous Driving with Large Language Models}},
  author={Wen, Licheng and Fu, Daocheng and Li, Xin and Cai, Xinyu and MA, Tao and Cai, Pinlong and Dou, Min and Shi, Botian and He, Liang and Qiao, Yu},
  booktitle={ICLR},
  year={2024}
}

@article{xu2024drivegpt4,
  title={{DriveGPT4: Interpretable End-to-end Autonomous Driving via Large Language Model}},
  author={Xu, Zhenhua and Zhang, Yujia and Xie, Enze and Zhao, Zhen and Guo, Yong and Wong, Kwan-Yee K and Li, Zhenguo and Zhao, Hengshuang},
  journal={IEEE Robotics and Automation Letters},
  year={2024},
  publisher={IEEE},
  	volume = {9},
	number = {10},
}

@article{cui2025vilad,
  title={{ViLaD: A Large Vision Language Diffusion Framework for End-to-End Autonomous Driving}},
  author={Cui, Can and Zhou, Yupeng and Peng, Juntong and Park, Sung-Yeon and Yang, Zichong and Sankaranarayanan, Prashanth and Zhang, Jiaru and Zhang, Ruqi and Wang, Ziran},
  journal={arXiv preprint arXiv:2508.12603},
  year={2025}
}

@inproceedings{nie2025large,
  title={{Large Language Diffusion Models}},
  author={Shen Nie and Fengqi Zhu and Zebin You and Xiaolu Zhang and Jingyang Ou and Jun Hu and JUN ZHOU and Yankai Lin and Ji-Rong Wen and Chongxuan Li},
  booktitle={NeurIPS},
  year={2025}
}

@article{chen2024end,
  title={{End-to-end Autonomous Driving: Challenges and Frontiers}},
  author={Chen, Li and Wu, Penghao and Chitta, Kashyap and Jaeger, Bernhard and Geiger, Andreas and Li, Hongyang},
  journal={IEEE Transactions on Pattern Analysis and Machine Intelligence},
  year={2024},
  publisher={IEEE},
  volume={46},
  number={12}
}

@inproceedings{chen2024drivinggpt,
  title={{DrivingGPT: Unifying Driving World Modeling and Planning with Multi-modal Autoregressive Transformers}},
  author={Chen, Yuntao and Wang, Yuqi and Zhang, Zhaoxiang},
  booktitle={ICCV},
  year={2025}
}

@article{huang2024drivegpt,
  title={{DriveGPT: Scaling Autoregressive Behavior Models for Driving}},
  author={Huang, Xin and Wolff, Eric M and Vernaza, Paul and Phan-Minh, Tung and Chen, Hongge and Hayden, David S and Edmonds, Mark and Pierce, Brian and Chen, Xinxin and Jacob, Pratik Elias and others},
  journal={arXiv preprint arXiv:2412.14415},
  year={2024}
}

@article{
hwang2024emma,
  title={{EMMA: End-to-End Multimodal Model for Autonomous Driving}},
author={Jyh-Jing Hwang and Runsheng Xu and Hubert Lin and Wei-Chih Hung and Jingwei Ji and Kristy Choi and Di Huang and Tong He and Paul Covington and Benjamin Sapp and Yin Zhou and James Guo and Dragomir Anguelov and Mingxing Tan},
journal={Transactions on Machine Learning Research},
issn={2835-8856},
year={2025},
}

@inproceedings{xing2025openemma,
  title={{OpenEMMA: Open-Source Multimodal Model for End-to-End Autonomous Driving}},
  author={Xing, Shuo and Qian, Chengyuan and Wang, Yuping and Hua, Hongyuan and Tian, Kexin and Zhou, Yang and Tu, Zhengzhong},
  booktitle={WACV Workshops},
  year={2025}
}

@inproceedings{shao_lmdrive_2023,
  title={{LMDrive: Closed-Loop End-to-End Driving with Large Language Models}},
  author={Shao, Hao and Hu, Yuxuan and Wang, Letian and Song, Guanglu and Waslander, Steven L and Liu, Yu and Li, Hongsheng},
  booktitle={CVPR},
  year={2024}
}

@article{wang_drivemlm_2023,
  title={{DriveMLM: Aligning Multi-Modal Large Language Models with Behavioral Planning States for Autonomous Driving}},
  author={Cui, Erfei and Wang, Wenhai and Li, Zhiqi and Xie, Jiangwei and Zou, Haoming and Deng, Hanming and Luo, Gen and Lu, Lewei and Zhu, Xizhou and Dai, Jifeng},
  journal={Visual Intelligence},
  volume={3},
  number={1},
  pages={22},
  year={2025},
  publisher={Springer}
}

@inproceedings{sima_drivelm_2024,
  title={{DriveLM: Driving with Graph Visual Question Answering}},
  author={Sima, Chonghao and Renz, Katrin and Chitta, Kashyap and Chen, Li and Zhang, Hanxue and Xie, Chengen and Bei{\ss}wenger, Jens and Luo, Ping and Geiger, Andreas and Li, Hongyang},
  booktitle={ECCV},
  year={2024},
}

@article{wang_drivecot_2024,
	title = {{DriveCoT}: {Integrating} {Chain}-of-{Thought} {Reasoning} with {End}-to-{End} {Driving}},
  author={Wang, Tianqi and Xie, Enze and Chu, Ruihang and Li, Zhenguo and Luo, Ping},
  journal={arXiv preprint arXiv:2403.16996},
  year={2024}
}

@inproceedings{caesar2020nuscenes,
  title={{nuScenes: A Multimodal Dataset for Autonomous Driving}},
  author={Caesar, Holger and Bankiti, Varun and Lang, Alex H and Vora, Sourabh and Liong, Venice Erin and Xu, Qiang and Krishnan, Anush and Pan, Yu and Baldan, Giancarlo and Beijbom, Oscar},
  booktitle={{CVPR}},
  year={2020}
}

@inproceedings{dinghint,
      title={{Hint-AD: Holistically Aligned Interpretability in End-to-End Autonomous Driving}},
      author={Ding, Kairui and Chen, Boyuan and Su, Yuchen and Gao, Huan-ang and Jin, Bu and Sima, Chonghao and Li, Xiaohui and Zhang, Wuqiang and Barsch, Paul and Li, Hongyang and others},
      booktitle={CoRL},
year={2024}
      }

@inproceedings{qian2024nuscenes,
  title={{NuScenes-QA: A Multi-Modal Visual Question Answering Benchmark for Autonomous Driving Scenario}},
  author={Qian, Tianwen and Chen, Jingjing and Zhuo, Linhai and Jiao, Yang and Jiang, Yu-Gang},
  booktitle={AAAI},
  year={2024}
}

@article{omeiza2021explanations,
  title={{Explanations in Autonomous Driving: A Survey}},
  author={Omeiza, Daniel and Webb, Helena and Jirotka, Marina and Kunze, Lars},
  journal={IEEE Transactions on Intelligent Transportation Systems},
  volume={23},
  number={8},
  pages={10142--10162},
  year={2021},
  publisher={IEEE}
}

@article{arrieta2020explainable,
  title={{Explainable Artificial Intelligence (XAI): Concepts, taxonomies, opportunities and challenges toward responsible AI}},
  author={Arrieta, Alejandro Barredo and D{\'\i}az-Rodr{\'\i}guez, Natalia and Del Ser, Javier and Bennetot, Adrien and Tabik, Siham and Barbado, Alberto and Garc{\'\i}a, Salvador and Gil-L{\'o}pez, Sergio and Molina, Daniel and Benjamins, Richard and others},
  journal={Information fusion},
  volume={58},
  pages={82--115},
  year={2020},
  publisher={Elsevier}
}

@inproceedings{ribeiro2016should,
  title={{'Why Should I Trust You?': Explaining the Predictions of Any Classifier}},
  author={Ribeiro, Marco Tulio and Singh, Sameer and Guestrin, Carlos},
  booktitle={SIGKDD},
  year={2016}
}

@inproceedings{lundberg2017unified,
  title={{A Unified Approach to Interpreting Model Predictions}},
  author={Lundberg, Scott M and Lee, Su-In},
  booktitle={NIPS},
  year={2017}
}

@inproceedings{selvaraju2017grad,
  title={{Grad-CAM: Visual Explanations From Deep Networks via Gradient-Based Localization}},
  author={Selvaraju, Ramprasaath R and Cogswell, Michael and Das, Abhishek and Vedantam, Ramakrishna and Parikh, Devi and Batra, Dhruv},
  booktitle={ICCV},
  year={2017}
}

@article{simonyan2013deep,
  title={{Deep Inside Convolutional Networks: Visualising Image Classification Models and Saliency Maps}},
  author={Simonyan, Karen and Vedaldi, Andrea and Zisserman, Andrew},
  journal={arXiv preprint arXiv:1312.6034},
  year={2013}
}

@inproceedings{kim2017interpretable,
  title={{Interpretable Learning for Self-Driving Cars by Visualizing Causal Attention}},
  author={Kim, Jinkyu and Canny, John},
  booktitle={ICCV},
  year={2017}
}

@inproceedings{kim2018textual,
  title={{Textual Explanations for Self-Driving Vehicles}},
  author={Kim, Jinkyu and Rohrbach, Anna and Darrell, Trevor and Canny, John and Akata, Zeynep},
  booktitle={ECCV},
  year={2018}
}

@inproceedings{kuhn2023textual,
  title={{Textual Explanations for Automated Commentary Driving}},
  author={K{\"u}hn, Marc Alexander and Omeiza, Daniel and Kunze, Lars},
  booktitle={IV},
  year={2023},
}

@inproceedings{jain2019attention, title={Attention is not Explanation}, author={Jain, Sarthak and Wallace, Byron C}, booktitle={NAACL}, year={2019} }

@inproceedings{qiang2022attcat,
  title={{AttCAT: Explaining Transformers via Attentive Class Activation Tokens}},
  author={Qiang, Yao and Pan, Deng and Li, Chengyin and Li, Xin and Jang, Rhongho and Zhu, Dongxiao},
  booktitle={NeurIPS},
  year={2022}
}

@inproceedings{xu2015show,
  title={{Show, Attend and Tell: Neural Image Caption Generation with Visual Attention}},
  author={Xu, Kelvin and Ba, Jimmy and Kiros, Ryan and Cho, Kyunghyun and Courville, Aaron and Salakhudinov, Ruslan and Zemel, Rich and Bengio, Yoshua},
  booktitle={ICML},
  year={2015},
}

@inproceedings{chen2024driving,
  title={{Driving with LLMs: Fusing Object-Level Vector Modality for Explainable Autonomous Driving}},
  author={Chen, Long and Sinavski, Oleg and H{\"u}nermann, Jan and Karnsund, Alice and Willmott, Andrew James and Birch, Danny and Maund, Daniel and Shotton, Jamie},
  booktitle={ICRA},
  year={2024},
}

@inproceedings{shrikumar2017learning,
  title={{Learning Important Features Through Propagating Activation Differences}},
  author={Shrikumar, Avanti and Greenside, Peyton and Kundaje, Anshul},
  booktitle={ICML},
  year={2017},
}

@INPROCEEDINGS{10919978,
  author={Cui, Can and Yang, Zichong and Zhou, Yupeng and Ma, Yunsheng and Lu, Juanwu and Li, Lingxi and Chen, Yaobin and Panchal, Jitesh and Wang, Ziran},
  booktitle={ITSC}, 
  title={{Personalized Autonomous Driving with Large Language Models: Field Experiments}}, 
  year={2024},
}

@article{cui2025llm4adlargelanguagemodels,
      title={{LLM4AD: Large Language Models for Autonomous Driving -- Concept, Review, Benchmark, Experiments, and Future Trends}}, 
      author={Can Cui and Yunsheng Ma and Sung-Yeon Park and Zichong Yang and Yupeng Zhou and Juanwu Lu and Juntong Peng and Jiaru Zhang and Ruqi Zhang and Lingxi Li and Yaobin Chen and Jitesh H. Panchal and Amr Abdelraouf and Rohit Gupta and Kyungtae Han and Ziran Wang},
      year={2025},
      journal={arXiv preprint arXiv:2410.15281},
}

@article{ma2025aln,
  title={{ALN-P3: Unified Language Alignment for Perception, Prediction, and Planning in Autonomous Driving}},
  author={Ma, Yunsheng and Yaman, Burhaneddin and Ye, Xin and Yurt, Mahmut and Luo, Jingru and Mallik, Abhirup and Wang, Ziran and Ren, Liu},
  journal={arXiv preprint arXiv:2505.15158},
  year={2025}
}

@article{qiao2025lightemma,
  title={{LightEMMA: Lightweight End-to-End Multimodal Model for Autonomous Driving}},
  author={Qiao, Zhijie and Li, Haowei and Cao, Zhong and Liu, Henry X},
  journal={arXiv preprint arXiv:2505.00284},
  year={2025}
}

@incollection{song2022anvil,
  title={{Anvil - System Architecture and Experiences from Deployment and Early User Operations}},
  author={Song, X Carol and Smith, Preston and Kalyanam, Rajesh and Zhu, Xiao and Adams, Eric and Colby, Kevin and Finnegan, Patrick and Gough, Erik and Hillery, Elizabett and Irvine, Rick and others},
  booktitle={PEARC},
  year={2022}
}

@incollection{boerner2023access,
  title={{ACCESS: Advancing Innovation: NSF’s Advanced Cyberinfrastructure Coordination Ecosystem: Services \& Support}},
  author={Boerner, Timothy J and Deems, Stephen and Furlani, Thomas R and Knuth, Shelley L and Towns, John},
  booktitle={PEARC},
  year={2023}
}

@inproceedings{jin2024tod3cap,
  title={{TOD3Cap: Towards 3D Dense Captioning in Outdoor Scenes}},
  author={Jin, Bu and Zheng, Yupeng and Li, Pengfei and Li, Weize and Zheng, Yuhang and Hu, Sujie and Liu, Xinyu and Zhu, Jinwei and Yan, Zhijie and Sun, Haiyang and others},
  booktitle={ECCV},
  year={2024},
}

@inproceedings{kimopenvla,
  title={{OpenVLA: An Open-Source Vision-Language-Action Model}},
  author={Kim, Moo Jin and Pertsch, Karl and Karamcheti, Siddharth and Xiao, Ted and Balakrishna, Ashwin and Nair, Suraj and Rafailov, Rafael and Foster, Ethan P and Sanketi, Pannag R and Vuong, Quan and others},
  booktitle={CoRL},
  year={2024},
}

@inproceedings{zitkovich2023rt,
  title={{RT-2: Vision-Language-Action Models Transfer Web Knowledge to Robotic Control}},
  author={Zitkovich, Brianna and Yu, Tianhe and Xu, Sichun and Xu, Peng and Xiao, Ted and Xia, Fei and Wu, Jialin and Wohlhart, Paul and Welker, Stefan and Wahid, Ayzaan and others},
  booktitle={CoRL},
  year={2023},
}

@article{tang2026hermes,
  title={{HERMES: A Holistic End-to-End Risk-Aware Multimodal Embodied System with Vision-Language Models for Long-Tail Autonomous Driving}},
  author={Tang, Weizhe and You, Junwei and Liu, Jiaxi and Wang, Zhaoyi and Gan, Rui and Huang, Zilin and Wei, Feng and Ran, Bin},
  journal={arXiv preprint arXiv:2602.00993},
  year={2026}
}

@inproceedings{hu2023planning,
  title={{Planning-Oriented Autonomous Driving}},
  author={Hu, Yihan and Yang, Jiazhi and Chen, Li and Li, Keyu and Sima, Chonghao and Zhu, Xizhou and Chai, Siqi and Du, Senyao and Lin, Tianwei and Wang, Wenhai and others},
  booktitle={CVPR},
  year={2023}
}

\end{document}